# A model for full local image interpretation

Guy Ben-Yosef[1] (guy.ben-yosef@weizmann.ac.il))
Liav Assif[1] (liav.assif@weizmann.ac.il)
Daniel Harari[1,2] (hararid@mit.edu)
Shimon Ullman[1,2] (shimon.ullman@weizmann.ac.il)
[1]Department of Computer Science, Weizmann Institute of Science, Rehovot, Israel
[2]Center for Brains, Minds and Machines, Massachusetts Institute of Technology, Cambridge, MA

**Abstract**

We describe a computational model of humans' ability to provide a detailed interpretation of a scene's components. Humans can identify in an image meaningful components almost everywhere, and identifying these components is an essential part of the visual process, and of understanding the surrounding scene and its potential meaning to the viewer. Detailed interpretation is beyond the scope of current models of visual recognition. Our model suggests that this is a fundamental limitation, related to the fact that existing models rely on feed-forward but limited top-down processing. In our model, a first recognition stage leads to the initial activation of class candidates, which is incomplete and with limited accuracy. This stage then triggers the application of class-specific interpretation and validation processes, which recover richer and more accurate interpretation of the visible scene. We discuss implications of the model for visual interpretation by humans and by computer vision models.

## Goal and introduction

Computational models of object recognition and categorization have made significant advances in recent years, demonstrating consistently improving results in recognizing thousands of natural object categories in complex natural scenes. However, in a number of key areas, existing models are far from approaching object recognition by the human visual system. A major limitation is the inability of current models to provide a detailed interpretation of a scene's components, which is an integral part of human recognition. Models may label for instance an image region as containing a horse, while humans looking at the image will naturally identify meaningful components almost everywhere, e.g. the right eye, the left ear, the mouth, mane, the right leg, the knee, the hoof, the tail, harness etc.

Identifying detailed components is an essential part of the visual process, leading to the understanding of the surrounding scene and its potential meaning to the viewer. However, interpretation is a difficult task since it requires the detection and localization of many semantic object parts, which can amount to dozens or even hundreds in a single image (Fig. 1A,B). By 'semantic' we mean components corresponding to object parts in the scene, such as 'tail' or 'tip of the ear', unlike 'curved contour' or 'dark region' describing image features. We approach the daunting problem of full accurate object interpretation by decomposing the full object or scene image into smaller, local, regions containing recognizable object components. As exemplified in Fig. 1B, in such local regions the task of full interpretation is still possible, but more tractable, since the number of semantic recognizable components is highly reduced. As will be shown, reducing the number of components plays a key factor in effective interpretation. At the same time, when the interpretation region becomes too limited, observers can no longer interpret or even identify its content, as illustrated in Fig. 1C. We therefore apply the interpretation process to local regions that are small, yet interpretable on their own by human observers.

The model proceeds by identifying within the local region a familiar configuration of semantic features learned from examples. This configuration is found by identifying the participating components, as well as their arrangement, which is defined by spatial relations among them. A central conclusion from the model is that full interpretation of even a local region at a human performance level depends on the use of relations that are currently not used by state-of-art, feed-forward image recognition models. These relations can be relatively complex, relying for example on computing local continuity, grouping and containment. We conclude from the model that the interpretation process is likely to be local and involve top-down processing. We propose a general scheme in which the interpretation process is applied initially to local and interpretable regions by combining bottom-up and top-down extraction of features and relations, and can subsequently be integrated and expanded to larger regions of interest.

The remaining of the paper proceeds as follows. In the next section we briefly review previous work related to image interpretation. Section 3 presents a model for full interpretation of local regions, with the goal of achieving interpretation at the level of detail obtained by humans in these regions. Section 4 describes experimental results to evaluate our model, and we conclude in Section 5 by discussing possible implications to our understanding of visual recognition and its mechanisms, and to the development of models and systems with visual capacities that are closer to human perceptual capacities.

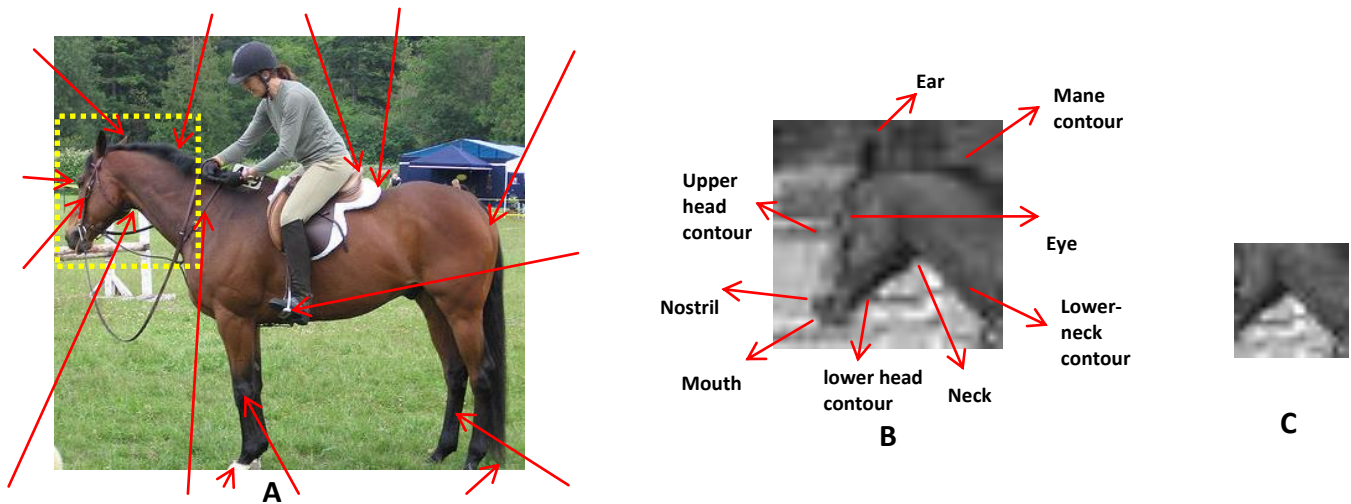

**Figure 1. (A).** A natural image in which humans can identify dozens of semantic features, arrows point to a subset of the identified features. **(B).** A local region and a set of features identified consistently by human observers; the number of semantic recognizable components is highly reduced. **(C).** When the local region becomes too limited, observers can no longer interpret or even identify its content.

## Image interpretation in previous work

Schemes related to object interpretation have been suggested under different names, including object parsing (e.g., Kokkinos & Yuille, 2009; Si & Zhu, 2013), or object grammar (e.g., Zhu et al., 2009), Fine-Grained recognition (e.g., Zhang et. al., 2014), and semantic segmentation (e.g., Chen. et al., 2015). Object parsing and object grammar models often refer to probabilistic frameworks that gradually decompose an object into simpler image structures. A recent representative work (Si & Zhu, 2013) describes a dictionary of object parts, represented by measures for shape, texture, and color. The parts primitives are learned in an unsupervised way and are not necessarily semantic in the sense described above. The relations between them are modeled by basic geometric structure indicating global and relative locations of the parts.

Some so called 'part-based models' provide another version of object interpretation at a coarser level (e.g., Deformable Part Model (DPM) (Felzenszwalb et al., 2010; Zhang et al., 2014). Such algorithms represent the object image by a set of part region primitives (e.g., HoG representation (Dalal & Triggs, 2005) or Convolutional Networks representation (Girshick et al.,2014)), which are learned in an unsupervised manner, and model basic geometric relations between them, such as relative location and distance. Such part-based models have proved highly useful for object recognition, however, the interpretation they provide is coarse and less localized compared with the current scheme (e.g. 'tail', 'wing' and 'body' for an airplane).

Several works have attempted to interpret a visual object by its contour features. Such schemes suggest a dictionary of informative contour fragments for the object, which is often learned in an unsupervised manner and often do not have a semantic meaning (Opelt et al., 2006; Arandjelovic & Zisserman, 2011; Ferrari et al., 2008). A more recent work (Hariharan et al., 2011) also suggested building a dictionary of semantic contours in a supervised manner, via human annotations of object images. The works above suggest several techniques to represent contours, and interpretation is obtained by matching contour in the image to contours in the dictionary, by modeling contour properties (e.g., curvature) and simple relations between contours, typically relative and global locations. Our approach extends such schemes by detecting points, contours, and regions of interest, that are semantic in the sense that humans can consistently recognize them in an image. As a result, we also use a significantly extended set of relations between the different types of feature primitives.

Fine-Grained recognition also aims to perform image interpretation by finding attributes and sub-category discrimination of the object in scene and its semantic parts. A recent example (Vedaldi et al. 2014) focuses on an aircraft benchmark. The scheme modeled aircrafts by a few semantic parts, e.g., the airplane nose, tail, wing, etc. and attributes of the plane or its parts such as 'does the plane have engine on tail?', or 'is it a propeller-plane?', etc.

Another form of image interpretation comes from work on so-called semantic segmentation, which attempts to make precise localization of the object surfaces in the scene. For example, a recent algorithm (Chen. et al 2015) based on features from the top layer of a 16-layers convolutional network, can identify the majority of pixels belonging to a horse surfaces in the PASCAL benchmark (Everingham et al., 2010), but it is far from predicting its precise boundary localization and detailed components. Our work differs from the approaches above since it aims to provide 'full' interpretation of the input image, namely, to localize all object parts that humans can interpret, and to learn to identify local configurations of these semantic features. We next turn to describe our interpretation scheme, and experiments done for its evaluation.

# Interpretation of local regions

Our goal is to describe the detailed semantic structure of a local region. More specifically, given a recognizable and interpretable local region in an object image, we aim to output the full semantic structure humans can find in this region. A natural choice for a formal description of semantic structures includes a set of primitive features and a set of relations defined over them. The primitive features are semantic components of the local region that are recognizable by observers (as in Fig. 1B). In a correct interpretation, the components are arranged in certain configurations, which are naturally defined by relations between components. The use of primitive components and relations between them is a common approach for modeling structured representations in areas of cognition and artificial intelligence (Russell & Norvig, 2005).

The semantic features to be identified by the model, e.g. 'ear', 'eye', 'neck', were supplied to the model using features which were consistently labeled in a prior experiment in Mechanical Turk (Crump et al., 2013). The ultimate task is then to identify these components in novel images by learning the image features and the relevant relations among them from examples. Our current model is not fully automatic, but relies on a set of spatial relations identified in previous works, and on an analysis of so-called 'hard negative' examples, described in the next section.

## Scheme overview

**Input: a local region to model** Our interpretation scheme begins by selecting a local recognizable object region, and getting from the Mechanical Turk a target set of semantic primitives to identify in it (e.g., Fig. 1B, Fig. 2A). The Mechanical Turk task required the naming of a certain highlighted object part. Consistent naming was examined and used to define the target interpretable components.

**Generating interpretation examples for learning** Next, we produced for learning a set of annotated images, in which the semantic features are marked manually (with automatic refinement). The final goal is to take a new image as an input, and mark in it all the detected semantic features (e.g., Figs. 2C,4,5). Having a set of positive interpretation examples, we next search for negative interpretation examples. The negative examples are collected by finding non-class images that are as similar as possible to true class instances. For this purpose, we trained a detector based on Bag of visual Words (Csurka et al., 2004) using the recent popular VLAD version (Jégou et al., 2010) with positive local region examples, and then applied these detectors to random images from known image benchmarks (PASCAL). The negative examples we use are non-class examples that received high detection scoring, and are therefore more confusable or 'hard' negatives for the detectors.

## Learning relations of correct interpretations

For each positive and negative example we compute a set of relations that exist between the annotated components. The relations are taken from a set of possible relations to compute (see below how this set is obtained). The relations identified in a given image are represented by a vector, where each component represents a specific relation. These vectors are then fed into a random forest classifier, which models the expected relations for correct primitive configurations in positive examples. We repeat this learning process for several iterations; at each iteration we add negative interpretation examples that obtained high scores in the previous iteration.

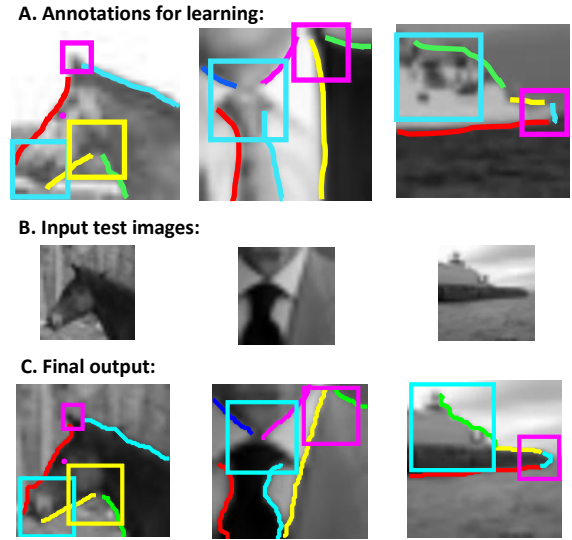

**Figure 2**. **(A).** Local recognizable images in which recognizable components are annotated. These components are the model primitives, which appear in three types: points, contours, and regions. From a set of annotated images, we learn the set of model relations. **(B).** Given a novel image, our target is to localize the set of primitives found in (A). **(C).** The interpretation scheme searches for combinations of primitives in the image, and output as interpretation the combination that matches best the learned set of relations.

**Interpretation of novel image** Given a raw novel image (e.g., Fig 2B), the scheme automatically searches for multiple combinations of image parts to serve as primitives, computes a relation vector for each combination, and by the learned classifier produces a score for the candidate combination. This search is feasible due to the small number of primitives in the local region. It finally returns the highest scoring combination as the interpretation of the input image (e.g., Fig. 2C).

## Primitives

To capture the recognized internal components fully as perceived by humans, our primitives are divided into three types, 2-D (regions), 1-D (contours), and 0-D (points). For example, a point-type primitive describes the eye in the horse head model (Fig. 2A, left panel), and a contour-type primitive describes borders such as the sides of the tie in the local region describing the man-in-suit model (Fig 2A, mid panel). Example sets of primitives for local regions are shown in Fig. 2A.

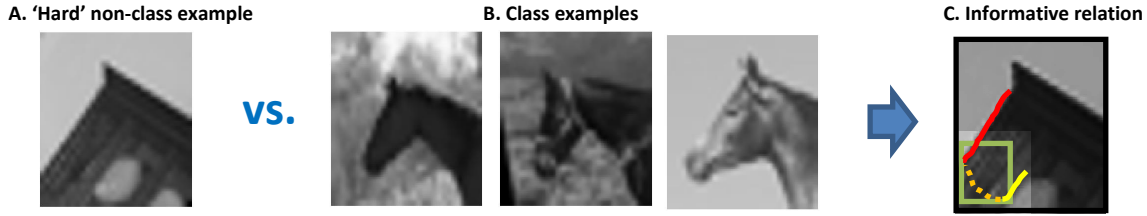

**Figure 3.** Inferring informative relations between internal components. **(A).** A 'hard' negative example, from Bag-of-Words classifier **(B).** Positive examples. We search for a relation between primitives that exists in the positive but not in the negative instance. An informative relation in this and other examples was contour 'bridging': the upper-head contour primitive (red), and the lower-head contour primitive (yellow) are linked by an edge through the mouth region primitive.

All three types of primitives have natural relations to both visual perception and computer vision models. Contour-type primitives have been linked to object perception from early studies (e.g., Attneave, 1954) and to explicit representation in the visual cortex (e.g., Pasupathy & Connor, 1999). Image contours are highly informative features that often correspond to meaningful object components. It proved difficult to learn and extract meaningful contours in complex images, and consequently contours turned less popular in recent recognition schemes, but in the current scheme they are more efficiently handled at the local region level. Point-primitive in our model include several types (in particular, high curvature points and local intensity extrema points), based on their use in both visual perception and computer vision (e.g., Attneave, 1954; Lindeberg, 1998; Lowe, 2004). Region-type descriptors proved highly efficient in computational models for identifying object regions under different viewing conditions (e.g., Dalal and Triggs, 2005, Felzenszwalb et al., 2010) and proved useful in the current model as well.

**Relations**

The set of relations used in our model was composed from two sources. One source consists of relations coming from prior computer and human vision modeling, such as proximity of points, contours and regions, or continuity and parallelism of contours. The second source includes relations inferred from our analysis of 'hard' non-class examples that were confused as positive by state-of-art detectors. More specifically, we have used the following iterative procedure:

1. Identify a 'hard' negative example that received high recognition score by region part detector based on Bag of visual Words, e.g., as in Fig 3A.

2. Identify a property or relation which exists in the positive set but not in the negative example (e.g., Fig 3C).

It is worth noting that identifying missing property or relations in step 2 becomes practical when analyzing small local regions, since the amount and complexity of primitives and relations is significantly reduced compared to standard object image. This learning process is in part manual; in human vision, it may come from a combination of evolutionary and learned components, see discussion. The relations coming from the second source include *cover* of a point by contour, *containment* of a contour or point in region, a contour that '*ends-in*' a region, and whether two disconnected contours can be '*bridged*' (i.e., linked by an edge in an edge map used by the model) consistent with the way they are connected in the positive image (see illustration in Fig. 3C). Our final library of relations includes unary relations (properties), binary relations, and relations among three or more primitives. Relations range from simpler interactions such as relative location, to more compound interactions such as continuity, containment, and bridging mentioned above.

**Experimental evaluation**

To evaluate our model and the library of derived relations, we performed experiments to assess the interpretation of novel images, by matching assignment of primitives to human annotations over multiple examples. To get positive examples, we randomly collected full-object images from known data sets (Flicker, Google images, ImageNet – Russakovsky et al., 2014), and then manually extracted from them the local region showing a particular object part for interpretation. We used local regions containing a horse head, a man in tie and suit, and a ship. These regions and the primitives defined for them are shown in Fig. 2. A large-scale experiment was done to evaluate the horse-head model, in which we collected 740 positive local image examples, from which we randomly selected 120 for training, and the rest used for testing. Negative set included 25000 images. Our experiments for the man-in-suit and ship local regions contained 60 positive examples, and 6000 negative examples.

To assess the extent of 'full' interpretation the model produces for novel images at a fine detail level, we manually annotated the semantic components recognized by human via Mechanical Turk for each tested positive example. We then automatically matched the ground truth annotated components to the interpretation output by correspondence criteria based on normalized Euclidean distance: for point, location distance; for contour, distances between ordered sample points; for regions, distance between centers.

Since our model is novel in terms of producing full interpretation, it cannot be compared directly in terms of completeness and accuracy with existing models. However, we made our set of annotations publically available and we provide baseline to match its results. Our results show an average matching error of 0.2442 normalized Euclidean distance over all eight primitive and 620 test images used for evaluating the horse head model. Example interpretation results for three models of Fig. 2 are presented in Fig. 4.A and in Fig. 5. Additional comparison measures can be used to assess full interpretation, which are left for future research.

To assess the role of complex relations, we compared our results to a version that uses the same interpretation scheme described above, but with a library containing unary and binary relations similar to those used in previous object interpretation schemes (as reviewed above), i.e., based on unary descriptions for shape and texture, and binary for relative location. In this reduced library we 'turned off' more complex relations from our analysis such as 'ends in' or 'bridging'. We show in Fig. 4A,B ten example pairs of the same image with two interpretations, full vs. reduced. Images were chosen randomly from our test set such that both schemes produced high interpretation score for them. Yet, the produced interpretations are perceptually different, and interpretation by the full-set scheme is significantly more precise. A comparison (not detailed here) shows that the fraction of primitives correctly localized by the full set scheme is increased by a factor of 1.45 than the reduced set version. An illustration of ten randomly selected images is shown in Fig. 4A,B.

## Discussion

**Local interpretation:** Results of the current study show that a detailed and well-localized interpretation can be obtained already from a limited image region, which contains a small number of elements, by using an appropriate set of relations. We suggest from the model and our experiments that efficient interpretation can start at the level of local regions, which can subsequently be integrated to produce more global interpretation of larger regions of interest.

**Top-scoring non-class detections:** of our model can be used in the future for two purposes. First, for validation: top-ranked false detections by bottom-up classification models often have low interpretation score (as in Fig. 4C), and therefore will be rejected by the interpretation stage. Second, we expect negative examples of high interpretation score to be perceptually similar to positive ones. We propose therefore to test psychophysically the agreement between human errors and errors made by models, with and without interpretation.

**A universal library of relations:** The set of relations needed for human-level interpretation is at present unknown. In this work we proposed a set starting from a collection of relations used in previous modeling as first approximation, and continued by adding relation candidates from analysis of hard non-class examples. This initial pool could be refined in the future by additional examples, leading ultimately to a universal set of useful interpretation relations. One finding of the current study is that simple spatial relations, such as displacements between primitives, are insufficient for a reliable interpretation. More complex relations, such as the 'bridgeability' of contours $c_i$, $c_j$, or a contour $c_i$ ending-in region $r_j$, contribute significantly to successful interpretation. In learning to interpret a new configuration, the set of candidate relations will be examined, and the informative ones for the task will be incorporated in the model.

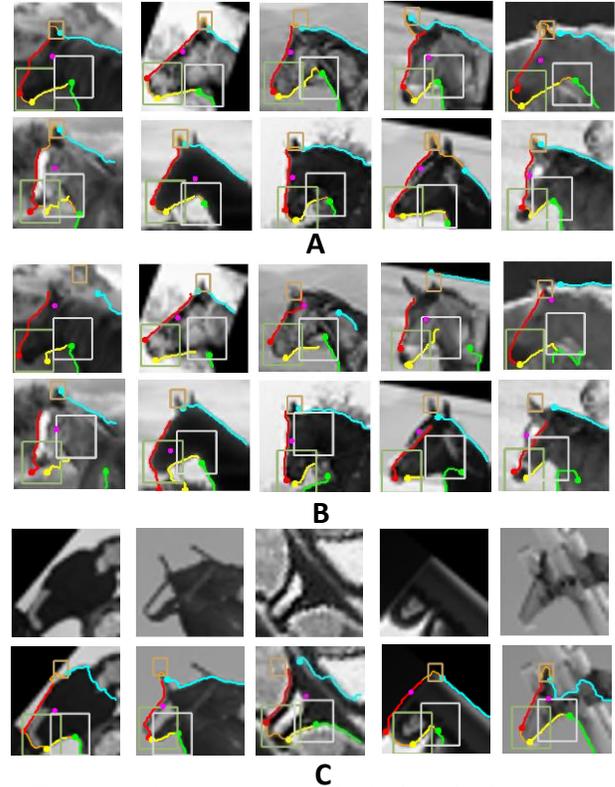

**Figure 4. (A).** Interpretation results for horse-head region for ten class examples. Here the scheme uses the full set of relations. **(B).** Interpretation results of the same images in (A), by a scheme using a reduced set of relation (see text for details). There are 6 mis-localization of primitives in (A) compared to 29 in (B). **(C).** Top-ranked Interpretation results for five non-class examples.

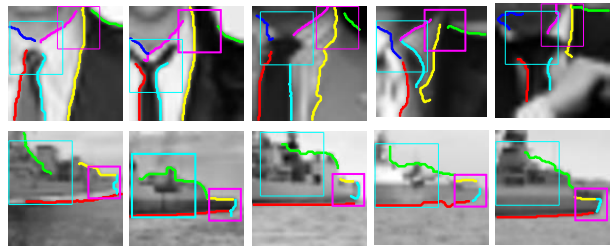

**Figure 5.** Examples of interpretation results for the man-in-suit and ship models described in Fig. 2.

**Implications for difficult visual tasks:** It will be interesting to examine in future studies the role of full interpretation in challenging visual tasks, which are beyond the scope of current computational theories, because they depend on fine localization of object parts and the relations between parts, as illustrated in Fig. 6. Full interpretation of components at the level produced by the current model is likely to prove useful for dealing with the interpretation of complex configurations arising in areas such as actions or social interactions between agents.

**Top-down processing:** Our model suggests that the relations required for a detailed interpretation are in part considerably more complex than spatial relations used in current recognition models. They are also often class-specific, in the sense that a relation such as 'connected by a smooth contour' is applied to a small selected set of components in some of the models. This suggests a scheme in which complex relations are computed at selected and class-specific locations. The recognition and interpretation process is naturally divided on this view to two main stages. The first is a bottom-up recognition stage, which may be similar to current high-performing computer vision models. This leads to the activation of objects models, which lacks detailed interpretation. Activated models will then trigger the application of top-down extraction of additional features and the computation of relevant relations to selected components, resulting in detailed interpretation as well as validation of the initial recognition stage.

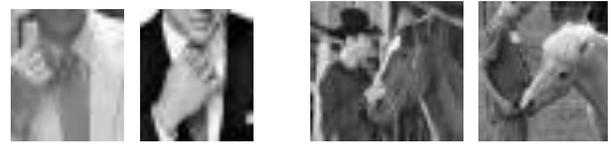

**Figure 6.** Full interpretation is useful for difficult visual tasks that are challenging for current computational theories. Such tasks include recognizing actions in local object-agent configurations, e.g., 'fixing a tie' (left pair), 'feeding a horse' (right pair) only one image in each pair is correct.